\documentclass{article}
\PassOptionsToPackage{numbers, compress}{natbib}

\usepackage[preprint]{neurips_2023}

\usepackage[utf8]{inputenc} 
\usepackage[T1]{fontenc}    
\usepackage{hyperref}       
\usepackage{multirow}
\usepackage{url}            
\usepackage{booktabs}       
\usepackage{amsfonts}       
\usepackage{amsmath}       
\usepackage{amsthm}
\usepackage{nicefrac}       
\usepackage{microtype}      
\usepackage{xcolor}         
\usepackage{csquotes}
\usepackage{graphicx}
\usepackage{wrapfig}

\newtheorem{thm}{Theorem}
\newtheorem{thm1}{Theorem1}
\newtheorem{principle}[thm]{Principle}
\newtheorem{corollary}[thm1]{Corollary}

\usepackage[capitalize]{cleveref}
\crefname{section}{Sec.}{Secs.}
\Crefname{section}{Section}{Sections}
\Crefname{table}{Table}{Tables}
\crefname{table}{Tab.}{Tabs.}

\title{Representation Learning in Anomaly Detection: Successes, Limits and a Grand Challenge}

\author{Yedid Hoshen\\ The Hebrew University of Jerusalem, Israel}

\begin{document}

\maketitle

\begin{abstract}
In this perspective paper, we argue that the dominant paradigm in anomaly detection cannot scale indefinitely and will eventually hit fundamental limits. This is due to the a no free lunch principle for anomaly detection. These limitations can be overcome when there are strong tasks priors, as is the case for many industrial tasks. When such priors do not exists, the task is much harder for anomaly detection. We pose two such tasks as grand challenges for anomaly detection: i) scientific discovery by anomaly detection ii) a "mini-grand" challenge of detecting the most anomalous image in the ImageNet dataset. We believe new anomaly detection tools and ideas would need to be developed to overcome these challenges. 
\end{abstract}

\section{Introduction}

Anomaly detection aims to discover rare, interesting phenomena. This task has been studied for many decades due to its central role for intelligent agents and its important industrial applications. These include: intrusion detection, video surveillance, industrial inspection, and preventive maintenance. Recently, the anomaly detection community has achieved remarkable progress, particularly in detecting anomalies in images. For example, the anomaly detection accuracy on the Cifar10 benchmark increased from around $65\%$ to nearly $99\%$ (ROC-AUC) over the last five years \cite{ocsvm, deepsvdd, zong2018deep, golan2018deep, hendrycks2019using, bergman2020classification, csi, droc, panda, mean_shifted, reiss_workshop}. This progress is mostly thanks to deep representation learning and incorporation of ever-improving foundation models. Indeed, this progress has nearly saturated some of the most popular datasets including the MVTec anomaly detection and segmentation benchmarks, both reaching near perfect accuracy.

The big question that arises is: \textit{what is next for anomaly detection?} This article aims to provide a possible direction. We will give a short, opinionated overview of representation-based anomaly detection. Then, we will demonstrate fundamental limits of this paradigm. Priors are a natural ways to overcome the impossibility results. This idea will be demonstrated on three different anomaly detection applications. Finally, we present a grand challenge which will require new methods for overcoming the limits of current methods while presenting rich rewards if successful.

\section{Representation Learning and the Promise of Foundation Models}

As mentioned in the introduction, anomaly detection seeks rare but interesting phenomena. These two objectives are quite different. Classical methods \cite{knn, breunig2000lof, gmm, ocsvm} have mostly dealt with the first objective, detecting rare sample as rarity can be well-defined mathematically. Specifically, provided a true probability density function of the data $p(x)$, rare items can be defined as those with low probability density. E.g., those $x \in X$ which have $p(x)$ smaller than some threshold. This presents a wealth of interesting and deep problems such as estimating $p(x)$ from a finite empirical sample, choosing a threshold and dealing with corrupted training samples. Note that the challenge of density estimation is a little different from standard density estimation in statistics and machine learning as the focus here is on the distribution tails rather than the most likely samples. A failure mode for such models is not properly estimating the probability density function due to sample complexity requirements. This commonly occurs when the data is high-dimensional and does not satisfy properties such as compactness or smoothness as is often the case for images represented as pixels.

A more fundamental issue with density estimation models is that some rare samples may not be semantically interesting. For example, minor variations in image background may make an image rare when performing density estimation in pixel space but may be of little interest. This is unfortunate as while low probability density is a well defined mathematical criterion ''interestingness'' is not. The objective of finding semantically interesting variation is also a goal of representation learning, which aims to map data into mathematical representations satisfying desired properties. Two properties that are particularly relevant to our task are: low-dimensionality and semantic similarity. The former specifies that learned representations should be of reduced dimensionality with respect to the original representation while preserving semantically important information. The latter specifies that samples with similar numerical values of their representations should also be similar in the degrees of variation which are semantically interesting. Density estimation in representation space therefore overcomes both limitations of classical methods. The low-dimensionality and smoother data manifold makes density estimation from a finite sample easier. As representation learning aims to focus on semantically interesting variation, rare values in representation space are likely to be more semantically meaningful.

A simple experiment performed by Bergman et al. \cite{bergman2020deep} tested the effectiveness of foundation model representations. The method used large ResNets trained on the ImageNet dataset to extract representations from all (train and test) images. The representations were simply the activations of the penultimate layer. The anomaly score for a test sample was defined as the average distance between the test image and K (~2) nearest neighbor training images, with distance measured in representation space. While very simple, this approach outperformed the state-of-the-art at the time. This performed well not only for datasets which are very similar to ImageNet such as Cifar10 but also for aerial and scientific images which are not found in ImageNet. Similar results were also shown by Perera and Patel \cite{perera2019learning}. Furthermore, Reiss et al. \cite{panda} found that finetuning the foundation model representation on the normal images provided in the training set achieved better results. A series of works performed the finetuning on: compactness \cite{panda}, contrastive \cite{mean_shifted} and DINO \cite{reiss_workshop} self-supervised objectives nearly saturating the popular Cifar10 benchmark. A similar idea \cite{cohen2020sub} can be used for anomaly segmentation, the task of detecting anomalies at the level of image pixel rather than entire images. Here too, foundation models are used to extract representations for every image region. E.g., by concatenating the spatial features of several ResNet or ViT blocks. Anomaly scoring can again be performed by density estimation. PatchCore \cite{patchcore}, a significant improvement of this approach, nearly saturated the popular MVTec anomaly segmentation benchmark \cite{mvtec2d}.

\section{No Free Lunch in Anomaly Detection}

The previous section may encourage the hypothesis that solving anomaly detection could simply be a matter of scaling up foundation models.  This is reasonable as in natural language processing a single foundation model was found to be competitive or better than nearly all benchmarks by scaling up unsupervised auto-regressive language models. This section uses the theoretical ideas from Reiss et al. \cite{reiss2023no} to argue that such scaling is unlikely to result in a single model for all anomaly detection tasks.

Let us a consider, as an example, a dataset consisting of images of birds (such as CUB200 \cite{cub200}). At training time, we observe many bird images that we deem normal. This dataset experiences significant variation in multiple attributes such as bird species, pose, background etc. At test time, we observe a new image which is sampled from a distribution combining the normal data and a set of anomalous birds with very long beaks. Assume that we have access to a perfect foundation model, which is able to craft representations according to our precise specifications. We would like to design the optimal representation for this task. The representation must include the attribute that differentiates between normal and anomalous data i.e. beak length. However, as we do not know the anomalies in advance, we generally cannot tell the FM to only output the beak length. Instead, we may choose a FM which outputs a concatenation of many attributes in the hope that the desired attribute will be among them. While such a representation may be sufficiently expressive, it will suffer from other issues. When the representation contains many attributes, the variation in the desired attribute between normal and anomalous data will be dwarfed by the variation due to noise in the other attributes. 

Reiss et al. \cite{reiss2023no} showed under a simplified Gaussian setting that the sensitivity of an anomaly detection algorithm (defined as the difference between the true and false positive rates) decreases with the number of attributes $d$ as $\frac{1}{\sqrt{d}}$. Increasing the number of attributes indefinitely will result in a random classifier. This presents an expressivity-sensitivity tradeoff, where the representation must be sufficiently expressive to include the (unspecified) desired attribute while not being too expressive and reducing sensitivity significantly. As there are (infinitely) many attributes, it is infeasible to guarantee a good result without priors or assumptions. This yields the no free lunch principle:

\begin{principle}
No Free Lunch in Anomaly Detection: A successful anomaly detection algorithm must choose the smallest set of attributes which include the (unspecified) anomalous attribute.
\end{principle}

\begin{corollary}
There is no one foundation model representation that fits all anomaly detection tasks
\end{corollary}

\section{Paying for Lunch with Strong Priors}

While the no free lunch principle may appear very restrictive, it can be overcome by introducing priors. These can be obtained quite naturally for many tasks. As their precise implementation is task specific, we present several examples illustrating how to do this in practice. 

\textbf{Point cloud AD.} In many practical situations e.g., autonomous driving or production line monitoring, 3D sensors (e.g. LIDAR) are used for measuring both the color and geometry of the scene. They often represent the data using point clouds, an unordered set with each element consisting of the (x,y,z) position and optionally the color (R,G,B) of the points. The task is to detect anomalies both at the level of the whole scene (“is this scene anomalous?”) and also anomaly localization (“which precise points are the anomalous ones?”). Horwitz and Hoshen \cite{horwitz2022empirical} first formulate our priors on the anomalies: i) anomalies occur in well localized regions. ii) they can occur in both color and geometry. iii) they are view invariant.  These priors are implemented by simply concatenating local view-invariant geometric representations e.g., longstanding handcrafted feature FPFH \cite{fpfh}, and local image features as done in SoTA method PatchCore \cite{patchcore}. This simple system outperformed carefully designed deep learning approaches e.g., \cite{bergmann2023anomaly} on a 3D anomaly dataset \cite{bergmann2021mvtec}. While both the image and geometric representations can be improved, these priors are unlikely to change and should guide future representation improvements. 

\textbf{Video AD.} Security camera monitoring is important for keeping important sites safe. As the number of cameras is growing faster than security officers automatic monitoring of noteworthy events is critical. This task has been researched by the anomaly detection community for decades. Our priors on this task, informed by current academic AD datasets, are that anomalies occur in object type, object speed and human pose (indicating unusual activities). Reiss and Hoshen \cite{ai_vad} therefore design a representation based on these priors by simply extracting optical flow (corresponding to object velocity), human pose (using an automated detector) and deep CLIP image encoder features which strongly correlate with object type. Density estimation is performed using standard techniques, kNN for high-dimensional features and GMM for low-dimensional features. This simple approach achieves the state-of-the-art performance on the three most popular unsupervised anomaly detection datasets \cite{ped2, avenue, shanghaitech}. This might indicate that more challenging benchmarks are required but the same approach would apply to future benchmarks including other anomaly types. 

\textbf{Logical AD.} So far, the anomalies we discussed either involved an entirely anomalous image, or having some localized anomalous elements in the image. Logical anomaly detection \cite{bergmann2022beyond} tackles the case where all image elements are normal on their own, but appear in an anomalous configuration. For example, in one dataset, each image has two screws, two washers and two nuts. An image with a nail instead of a screw would be the typical structural anomaly (as nails are and anomalous element). On the other hand, an image with two screws, one washer and three nuts is a logical anomaly as the elements are normal but their configuration is not. One prior we can use is that the ordering between elements does not matter but only their distribution does. Guided by this prior, Cohen et al. \cite{cohen2023set} proposed a deep set representation for image regions, related to the Sliced Wasserstein distance. The representation is quite simple to implement. It requires projecting the representation of each element along $p$ randomly sampled 1D directions (we used $p=500$). A histogram is constructed to summarize the projected values along each direction for every image. The image is represented by the concatenation of the histograms of all directions. Density estimation is performed using kNN (after a prior whitening stage). This simple approach obtains the best overall accuracy on the MVTec-LOCO dataset \cite{bergmann2022beyond}. 

To summarize, we presented three tasks, each required task specific priors. The priors guided the desired representations, and were easy to implement in practice. Anomaly scoring was a standard process for all tasks. Given good priors, these approaches achieved higher accuracy than more complex approaches. This simple paradigm is easily extended to other anomaly detection tasks, particularly in industrial settings where practitioners have experience with their systems and their failure modes, but where labelled anomalies are unavailable due to their rarity or high variability.

\section{A Grand Challenge}

We propose anomaly detection for scientific discovery as a grand challenge for anomaly detection. Anomalies are an essential part of the scientific discovery process. This was formulated by Thomas Kuhn, a prominent philosopher of science, in his magnum opus “The Structure of Scientific Revolutions” \cite{kuhn2012structure}. He established that “discovery begins with the awareness of anomaly” and that anomaly detection is essential for facilitating scientific paradigm shift. As an illustrative example, we can recall Fleming’s discovery of Penicillin. Upon returning from vacation, Fleming inspected his Petri dishes which were left when he was away. He noticed that while they had bacteria growing in them, patches containing mold were free of bacteria. This anomalous behavior prompted him to inspect the mold more carefully eventually leading to the discovery of penicillin, a cornerstone of modern medicine. Admittedly, not all scientific anomalies are an obvious fit for a computational approach, as some require designing specialized experiments in order to detect them e.g. the Michelson-Morley experiment. However, there are important disciplines with vast amounts of data e.g. cell biology, -omics, particle physics, astrophysics, climate, to name but a few.

As a preliminary exploration, Michael et al. \cite{Michael-Pitschaze2023.04.03.535457} tackled the task of detecting anomalous proteins. They created a training set consisting of human proteins (normal), and a test set consisting of human proteins (normal) and virus proteins (anomalous), representing each proteins using a deep auto-regressive encoder pretrained on a vast dataset (ESM \cite{rives2019biological}). Anomaly scoring was performed using kNN. The method was able to detect virus proteins at $98.7\%$ ROC-AUC. Such a method may be able to detect intrusion of unknown proteins into the body. 

Scientific anomaly detection is much harder than the tasks that current methods are able to tackle. The primary reason is the lack of good priors on selecting the attributes along which the next discovery will emerge. Differently from industrial tasks where we can confidently devise such priors, scientific discoveries are often entirely unexpected. Devising effective ways of developing such priors is the key for this grand challenge. Several other issues include: i) scientific data are often comprised of multiple data modalities (lab notes, imagery, spectroscopy etc.), whereas most current AD methods operate only at the level of a single modality. ii) the data is unsupervised, there is no training set consisting only of normal samples iii) a fundamental point is that anomalies in science often mean a sample that deviates from the scientific paradigm rather than a low-likelihood sample. Incorporating the current paradigm into the anomaly detection algorithms is an interesting challenge for the future.

As a stepping stone, we propose a “less grand” but difficult challenge. We pose the deceptively simple but fundamental question:

\begin{center}
\textit{What is the most anomalous image on ImageNet?}
\end{center}

This question shares some of the challenges of scientific AD. We do not have good priors on which attributes are the most relevant ones. The data are unsupervised, we are not provided with a notion of normality. However the task is simpler from scientific AD in the sense that there are very powerful FMs for images and the data are unimodal. Solving this task, which is closer to previous ones tackled by the community is likely to provide a good stepping stone on the way to the grand challenge. 

To summarize, scientific anomaly detection holds many fundamental challenges but also may make computational anomaly detection an essential tool for scientific discovery. This will be a grand challenge for anomaly detection for the foreseeable future.

\section{Conclusion}

Foundation models were a major component in the advances of anomaly detection over the past few years. However, we argued that scaling them up will not be able to improve anomaly detection indefinitely, due to the no-free-lunch principle.  A successful current paradigm which overcomes the no-free-lunch principle is to assume task priors over the anomalies. While this is a valid approach for many industrial tasks, it is powerless when we do not know of such priors. We therefore proposed anomaly detection for scientific discovery as a grand challenge for anomaly detection and unsupervised anomaly detection on ImageNet as a mini-grand challenge.

{\small

\bibliographystyle{unsrt}
\bibliography{cite}

\begin{thebibliography}{10}

\bibitem{ocsvm}
Bernhard Scholkopf, Robert~C Williamson, Alex~J Smola, John Shawe-Taylor, and
  John~C Platt.
\newblock Support vector method for novelty detection.
\newblock In {\em NIPS}, 2000.

\bibitem{deepsvdd}
Lukas Ruff, Nico Gornitz, Lucas Deecke, Shoaib~Ahmed Siddiqui, Robert
  Vandermeulen, Alexander Binder, Emmanuel M{\"u}ller, and Marius Kloft.
\newblock Deep one-class classification.
\newblock In {\em ICML}, 2018.

\bibitem{zong2018deep}
Bo~Zong, Qi~Song, Martin~Renqiang Min, Wei Cheng, Cristian Lumezanu, Daeki Cho,
  and Haifeng Chen.
\newblock Deep autoencoding gaussian mixture model for unsupervised anomaly
  detection.
\newblock In {\em International conference on learning representations}, 2018.

\bibitem{golan2018deep}
Izhak Golan and Ran El-Yaniv.
\newblock Deep anomaly detection using geometric transformations.
\newblock In {\em NeurIPS}, 2018.

\bibitem{hendrycks2019using}
Dan Hendrycks, Mantas Mazeika, Saurav Kadavath, and Dawn Song.
\newblock Using self-supervised learning can improve model robustness and
  uncertainty.
\newblock In {\em NeurIPS}, 2019.

\bibitem{bergman2020classification}
Liron Bergman and Yedid Hoshen.
\newblock Classification-based anomaly detection for general data.
\newblock In {\em ICLR}, 2020.

\bibitem{csi}
Jihoon Tack, Sangwoo Mo, Jongheon Jeong, and Jinwoo Shin.
\newblock Csi: Novelty detection via contrastive learning on distributionally
  shifted instances.
\newblock {\em NeurIPS}, 2020.

\bibitem{droc}
Kihyuk Sohn, Chun-Liang Li, Jinsung Yoon, Minho Jin, and Tomas Pfister.
\newblock Learning and evaluating representations for deep one-class
  classification.
\newblock {\em arXiv preprint arXiv:2011.02578}, 2020.

\bibitem{panda}
Tal Reiss, Niv Cohen, Liron Bergman, and Yedid Hoshen.
\newblock Panda: Adapting pretrained features for anomaly detection and
  segmentation.
\newblock In {\em Proceedings of the IEEE/CVF Conference on Computer Vision and
  Pattern Recognition}, pages 2806--2814, 2021.

\bibitem{mean_shifted}
Tal Reiss and Yedid Hoshen.
\newblock Mean-shifted contrastive loss for anomaly detection.
\newblock {\em arXiv preprint arXiv:2106.03844}, 2021.

\bibitem{reiss_workshop}
Tal Reiss, Niv Cohen, Eliahu Horwitz, Ron Abutbul, and Yedid Hoshen.
\newblock Anomaly detection requires better representations.
\newblock In {\em Computer Vision--ECCV 2022 Workshops: Tel Aviv, Israel,
  October 23--27, 2022, Proceedings, Part IV}, pages 56--68. Springer, 2023.

\bibitem{knn}
Eleazar Eskin, Andrew Arnold, Michael Prerau, Leonid Portnoy, and Sal Stolfo.
\newblock A geometric framework for unsupervised anomaly detection.
\newblock In {\em Applications of data mining in computer security}, pages
  77--101. Springer, 2002.

\bibitem{breunig2000lof}
Markus~M Breunig, Hans-Peter Kriegel, Raymond~T Ng, and J{\"o}rg Sander.
\newblock Lof: identifying density-based local outliers.
\newblock In {\em ACM sigmod record}, volume~29, pages 93--104. ACM, 2000.

\bibitem{gmm}
Michael Glodek, Martin Schels, and Friedhelm Schwenker.
\newblock Ensemble gaussian mixture models for probability density estimation.
\newblock {\em Computational Statistics}, 28(1):127--138, 2013.

\bibitem{bergman2020deep}
Liron Bergman, Niv Cohen, and Yedid Hoshen.
\newblock Deep nearest neighbor anomaly detection.
\newblock {\em arXiv preprint arXiv:2002.10445}, 2020.

\bibitem{perera2019learning}
Pramuditha Perera and Vishal~M Patel.
\newblock Learning deep features for one-class classification.
\newblock {\em IEEE Transactions on Image Processing}, 28(11):5450--5463, 2019.

\bibitem{cohen2020sub}
Niv Cohen and Yedid Hoshen.
\newblock Sub-image anomaly detection with deep pyramid correspondences.
\newblock {\em arXiv preprint arXiv:2005.02357}, 2020.

\bibitem{patchcore}
Karsten Roth, Latha Pemula, Joaquin Zepeda, Bernhard Sch{\"o}lkopf, Thomas
  Brox, and Peter Gehler.
\newblock Towards total recall in industrial anomaly detection.
\newblock In {\em Proceedings of the IEEE/CVF Conference on Computer Vision and
  Pattern Recognition}, pages 14318--14328, 2022.

\bibitem{mvtec2d}
Paul Bergmann, Michael Fauser, David Sattlegger, and Carsten Steger.
\newblock Mvtec ad--a comprehensive real-world dataset for unsupervised anomaly
  detection.
\newblock In {\em Proceedings of the IEEE/CVF conference on computer vision and
  pattern recognition}, pages 9592--9600, 2019.

\bibitem{reiss2023no}
Tal Reiss, Niv Cohen, and Yedid Hoshen.
\newblock No free lunch: The hazards of over-expressive representations in
  anomaly detection.
\newblock {\em arXiv preprint arXiv:2306.07284}, 2023.

\bibitem{cub200}
Catherine Wah, Steve Branson, Peter Welinder, Pietro Perona, and Serge
  Belongie.
\newblock The caltech-ucsd birds-200-2011 dataset.
\newblock 2011.

\bibitem{horwitz2022empirical}
Eliahu Horwitz and Yedid Hoshen.
\newblock An empirical investigation of 3d anomaly detection and segmentation.
\newblock {\em arXiv preprint arXiv:2203.05550}, 2022.

\bibitem{fpfh}
Radu~Bogdan Rusu, Nico Blodow, and Michael Beetz.
\newblock Fast point feature histograms (fpfh) for 3d registration.
\newblock In {\em 2009 IEEE International Conference on Robotics and
  Automation}, pages 3212--3217, 2009.

\bibitem{bergmann2023anomaly}
Paul Bergmann and David Sattlegger.
\newblock Anomaly detection in 3d point clouds using deep geometric
  descriptors.
\newblock In {\em Proceedings of the IEEE/CVF Winter Conference on Applications
  of Computer Vision}, pages 2613--2623, 2023.

\bibitem{bergmann2021mvtec}
Paul Bergmann, Xin Jin, David Sattlegger, and Carsten Steger.
\newblock The mvtec 3d-ad dataset for unsupervised 3d anomaly detection and
  localization.
\newblock {\em arXiv preprint arXiv:2112.09045}, 2021.

\bibitem{ai_vad}
Tal Reiss and Yedid Hoshen.
\newblock Attribute-based representations for accurate and interpretable video
  anomaly detection.
\newblock {\em arXiv preprint arXiv:2212.00789}, 2022.

\bibitem{ped2}
Vijay Mahadevan, Weixin Li, Viral Bhalodia, and Nuno Vasconcelos.
\newblock Anomaly detection in crowded scenes.
\newblock In {\em 2010 IEEE computer society conference on computer vision and
  pattern recognition}, pages 1975--1981. IEEE, 2010.

\bibitem{avenue}
Cewu Lu, Jianping Shi, and Jiaya Jia.
\newblock Abnormal event detection at 150 fps in matlab.
\newblock In {\em Proceedings of the IEEE international conference on computer
  vision}, pages 2720--2727, 2013.

\bibitem{shanghaitech}
Wen Liu, Weixin Luo, Dongze Lian, and Shenghua Gao.
\newblock Future frame prediction for anomaly detection--a new baseline.
\newblock In {\em Proceedings of the IEEE conference on computer vision and
  pattern recognition}, pages 6536--6545, 2018.

\bibitem{bergmann2022beyond}
Paul Bergmann, Kilian Batzner, Michael Fauser, David Sattlegger, and Carsten
  Steger.
\newblock Beyond dents and scratches: Logical constraints in unsupervised
  anomaly detection and localization.
\newblock {\em International Journal of Computer Vision}, 130(4):947--969,
  2022.

\bibitem{cohen2023set}
Niv Cohen, Issar Tzachor, and Yedid Hoshen.
\newblock Set features for fine-grained anomaly detection.
\newblock 2023.

\bibitem{kuhn2012structure}
Thomas~S Kuhn.
\newblock {\em The structure of scientific revolutions}.
\newblock University of Chicago press, 1962.

\bibitem{Michael-Pitschaze2023.04.03.535457}
Tomer Michael-Pitschaze, Niv Cohen, Dan Ofer, Yedid Hoshen, and Michal Linial.
\newblock Detecting anomalous proteins using deep representations.
\newblock {\em bioRxiv}, 2023.

\bibitem{rives2019biological}
Alexander Rives, Joshua Meier, Tom Sercu, Siddharth Goyal, Zeming Lin, Jason
  Liu, Demi Guo, Myle Ott, C.~Lawrence Zitnick, Jerry Ma, and Rob Fergus.
\newblock Biological structure and function emerge from scaling unsupervised
  learning to 250 million protein sequences.
\newblock {\em PNAS}, 2019.

\end{thebibliography}
}

\end{document}